\documentclass{article}

\usepackage{PRIMEarxiv}

\usepackage[utf8]{inputenc} 
\usepackage[T1]{fontenc}    
\usepackage{hyperref}       
\usepackage{url}            
\usepackage{booktabs}       
\usepackage{amsfonts}       
\usepackage{nicefrac}       
\usepackage{microtype}      
\usepackage{lipsum}
\usepackage{fancyhdr}       
\usepackage{graphicx}       
\graphicspath{{media/}}     
\usepackage{amsmath}
\DeclareMathOperator{\atantwo}{atan2}
\DeclareMathOperator{\atan}{atan}
\newcommand{\bea}{\begin{eqnarray}}
\newcommand{\eea}{\end{eqnarray}}
\newcommand{\comment}[1]{}
\newcommand{\ii}{\boldsymbol{i}}
\newcommand{\jj}{\boldsymbol{j}}
\newcommand{\kk}{\boldsymbol{k}}

\newcommand{\HH}{\mathbf{H}}

\newcommand{\abs}[1]{|#1|}
\title{A trainable monogenic  ConvNet layer robust in front of large contrast changes in image classification
\thanks{\textit{Under review}}
}

\author{
  E. Ulises Moya-Sánchez, \\
  Jalisco Gov /Universidad Autónoma de Guadalajara \\
  Guadalajara, Jalisco, México\\
  \texttt{eduardo.moya\{@jalisco.gob.mx, @edu.uag.mx\}} \\
   \And
  Sebastiá Xambo-Descamps \\
  Barcelona Tech and  Barcelona Supercomputing Center \\
  Barcelona, Spain\\
  \texttt{sebastia.xambo@upc.edu} \\
   \And
 Abraham Sánchez \\
  Jalisco Gov \\
  Guadalajara, Jalisco, Mexico\\
  \texttt{abraham.sanchez@jalisco.gob.mx} \\
  \And
  Sebastian Salazar-Colores \\
  Centro de Investigaciones en Óptica \\
  León, Guanajuato, Mexico\\
  \texttt{sebastian.salazar@cio.mx} \\
   \And
 
  Ulises Cortés \\
  Barcelona Tech and  Barcelona Supercomputing Center \\
  Barcelona, Spain\\
  \texttt{Ulises.Cortes@bsc.es} \\
}

\begin{document}
\maketitle

\begin{abstract}
Convolutional Neural Networks (ConvNets) at present achieve remarkable performance in image classification tasks.  However, current ConvNets cannot guarantee the capabilities of the mammalian visual systems such as invariance to contrast and illumination changes.  Some ideas to overcome the illumination and contrast variations usually have to be tuned manually and tend to fail when tested with other types of data degradation. In this context,  we present a new bio-inspired {entry} layer, M6, which detects low-level geometric features (lines, edges, and orientations) which are similar to patterns detected by the V1 visual cortex. This new trainable layer is capable of coping with image classification even with large contrast variations.  The explanation for this behavior is the monogenic signal geometry, {which represents each pixel value in a 3D space using quaternions}, a fact that confers a degree of explainability to the networks. We compare M6 with a conventional convolutional layer (C) and a deterministic quaternion local phase layer (Q9). The experimental setup {is designed to evaluate the robustness} of our M6 enriched ConvNet model and  includes three architectures, four datasets, three types of contrast degradation (including non-uniform haze degradations). The numerical results reveal that the models with M6 are the most robust in front of any kind of contrast variations. This amounts to a significant enhancement of the C models, which usually have reasonably good performance only when the same training and test degradation are used, except for the case of maximum degradation.
Moreover, the  Structural Similarity Index Measure (SSIM)  is used to analyze and explain the robustness effect of the M6 feature maps under any kind of contrast degradations.
\end{abstract}


\section{Introduction}

{O}{ne} of of the major features of the mammalian visual cortex is its built-in capacity to  recognize objects  independently of size, contrast, illumination,  angle of view, or brightness, among other quantities \cite{rolls-stringer-2006}. Achieving an equivariance or invariance response to these transformations is an important goal in Deep Learning (DL) \cite{hendrycks-dietterich-2019,dodge-karam-2016}. In fact, large contrast changes, such as the contrast of day and night, or as it appears in different sorts of glare, could be a problem in many DL applications such as self-driving cars \cite{wang-yao-2018}, surface glazes in medical images \cite{halicek-febelo-fei-2019}, or day-round surveillance, among others.

Among the ideas to overcome the illumination and the contrast variations in image classification problems, we find data augmentation~\cite{simard-steinkraus-platt-2003,hernandez-konig-2018}.  However, previous work indicates that  learning of an invariant response may fail even with large data augmentations in the training process \cite{cohen-welling-2016}. In addition,  the data augmentation techniques have three main problems~\cite{ratner-ehrenberg-etal-2017, dao-gu-ratner-etal-2019}:   i) it has to be  tuned (manually) by  human experts and in practice this causes large variances in the DL-model performance, ii) because of a lack of analytic tools (even for simple models), it is not well-understood how training on augmented data affects the learning process, iii) data augmentation approaches focus on improving the overall performance of a model, and it is often imperative to have a finer-grained perspective. This means that data augmentation techniques are required to mitigate the weak performance in dealing with the under-represented classes.

Another common idea to tackle the contrast and illumination problems is the data normalization, such as local response normalization~\cite{krizhevsky-sutskever-hinton-2012}. However, the main problem of these approaches appears when the change of illumination is non-uniform across the images in the data set~\cite{rad-roth-lepetit-2020}. {Rad et al.  proposed to use an adaptive local contrast normalization based on a window (region) difference of Gaussians for image detection. Although their approach is novel, the parameters of the Gaussian functions are based on the dataset illumination. As a result, for a new image with very different illumination  the detection performance in general will be reduced.}

The usage of Generative Adversarial Networks (GAN)  is capable of restoring  contrast affected images~\cite{hendrycks-dietterich-2019,fan-yang-2017}. Nevertheless, these architectures  are primarily designed for visual restoration, not for image classification. 

The limitations of these  approaches reveal  opportunities to explore novel methods. In this work, we propose a new strategy to progress toward achieving contrast-illumination invariance in image classification tasks.  Specifically, we present a new bio-inspired trainable layer, M6, which detects low-level geometric features (such as lines, edges and orientations) which are not unlike similar patterns detected by the V1 visual cortex \cite{hubel-wiesel-1962}.

The M6 layer is based on the 2D extension of the analytic signal, called \textit{monogenic signal}. As a result, each pixel value of an image $I(x,y)$  $\in \mathbf{R}$ is mapped to a Hamilton quaternion $\mathbf{H}$ (see Figure \ref{fig:3D_mono}). The geometry of this approach has {the} remarkable simple property that the quality of the representation is not affected by large changes of the pixel intensities, a fact that confers a degree of explainability to the networks.

On the experimental side, to evaluate the predictive performance and robustness of M6, we have simulated contrast changes in three different ways using four datasets (MNIST \cite{lecun-bottou-orr-muller-1998}, Fashion MNIST \cite{han-kashif-2017}, CIFAR-10 \cite{krizhevsky-Hinton-etothers-2009}, and Dogs and Cats \cite{elson-doucer-2007}) with three architectures: A1 (shallow), A2 (medium), and A3 (very deep). We compare the performance of M6 against a conventional ConvNets layer (C) and the Q9 layer~\cite{moya-xambo-perez-salazar-mzortega-cortes-2020}, which follows a close approach to handle contrast changes.
To evaluate the robustness response the models were trained with a specific data degradation and tested with other type of data degradations. The numerical results in the test set confirm that M6 achieves a remarkable resilient response to contrast variations when compared to standard convolution networks and with the Q9 approach. 


The rest of the paper is organized  as follows: In Section \ref{sec:related}, previous related works are acknowledged. Section  \ref{sec:backg} summarizes background materials concerning the local phase computation, the monogenic signal, and bio-inspired tools. In section \ref{sec:mono}, we present the computation aspects of the monogenic layer M6, and in Section \ref{sec:exp_setup} the data and experimental  arrangements. The  experimental outcomes and their analysis are described in Section \ref{sec:results},  and the conclusions and future work in Section \ref{sec:conl}.

\section{Related work} \label{sec:related}

Promising approaches were advanced in the bio-inspired papers \cite{rolls-stringer-2006} (which introduces \textbf{VisNet}) and \cite{leibo-liao-2015}. They introduce and study, via quite different approaches, interesting hypothesis about how the cortex achieves various invariant representations. However, testing is done on relatively small datasets and although lighting invariance is considered, contrast invariance is not. 

Now let us mention the works \cite{worrall-garbin-daniyar-brostow-2017} and \cite{richards-paiement-xie-duc-2020}. The mathematical tools used in these papers involve only complex numbers, but otherwise are somewhat akin to the ones used here. Their goals, however, are quite different and in particular they do not seek a robust response to large contrast or illumination changes. The first presents (complex) harmonic networks (\textbf{H-nets}) and shows, by means of complex circular harmonic functions, that they exhibit equivariance to patch-wise translations and to in-plane rotations. The aim of the second is to present a complex-valued learnable Gabor-modulated network that features orientation robustness. 

Recent developments dealing with contrast robustness have been reported. In \cite{dapello-marques-schrimpf-geiger-cox-dicarlo-2020}, for instance, the authors  present an entry layer (\textbf{VOneNet}) which uses Gabor functions and report the results of testing it for different data degradations and perturbations. The layer is deterministic (non-trainable) and consequently it requires fine tuning of several parameters (related to the input size, normalization, and random parameters) to set the Gabor function. For contrast degradation, the test accuracy drops form 75.6\% to 28.5\% on the Imagenet dataset.

Another deterministic approach is presented in \cite{alshammari-akcay-breckon-2018}.  The authors propose to combine the chromatic component of a perceptual color-space to improve image segmentation. Nevertheless, tests in this work deal only with outdoor images with normal degradations and their parameters have to be tuned according to the dataset. 

A previous version of the M6 layer, still deterministic, was presented in \cite{moya-xambo-salazar-sanchez-cortes-2021}, but already with quite robust responses for illumination or contrast variations. In another recent work~\cite{moya-xambo-perez-salazar-mzortega-cortes-2020}, the authors propose a bio-inspired approach to leverage quaternion Gabor filters to improve the classification even with contrast degradations. An adaptive local contrast normalization was proposed in \cite{rad-roth-lepetit-2020}. Although their approach is more effective than conventional normalizations, the trainable parameters are based on the dataset illumination. As a result, for images with different degradations the detection performance will in general be reduced.

To assess in more concrete terms the performance and robustness of our M6 layer, we primarily compare it to conventional convolutional layers C, which are the most popular and are the architectures of many ConvNet models. Because of that, we looked for papers satisfying the following conditions: i) robust to contrast or illumination changes, ii) code available in Tensorflow, iii) implementable in different architectures and datasets, iv) applicable for image classification. The selected papers for that purpose, already cited above, are
\cite{dapello-marques-schrimpf-geiger-cox-dicarlo-2020,alshammari-akcay-breckon-2018,rad-roth-lepetit-2020,moya-xambo-perez-salazar-mzortega-cortes-2020}, and on closer inspection we find that only \cite{moya-xambo-perez-salazar-mzortega-cortes-2020} satisfies the four criteria. As we mention before, the main strength of M6 with respect to \cite{moya-xambo-perez-salazar-mzortega-cortes-2020} is that the latter is based on a deterministic unit, while M6 is trainable, a fact that is responsible for all its goods reported in this work.

\section{Background}\label{sec:backg}

In this section we explain background notions and notations used in this work. These include the properties of the analytic and monogenic signals. Finally, we explain the bio-inspired connection with the proposed layer. 

\subsection{Signals}
We define 1D (resp. 2D) \textit{multivectorial signals}  as $C^1$ maps $U \to \mathcal{G}$  from an interval $U \subset \mathbf{R}$ (a region $ U \subset \mathbf{R}^2$) into a \textit{geometric algebra} $\mathcal{G}$ (see \cite{xambo-2018-spins} for detalied definition).   For $\mathcal{G}=\mathbf{R}$  ($\mathcal{G}=\mathbf{C}$,  $\mathcal{G}=\mathbf{H}$)  we say that the signal is \textit{scalar} (\textit{complex, quaternionic}). For technical reasons, we also assume that signals are in $L^2$ (that is, their modulus is square-integrable). 

\subsection{Analytic signal} \label{ssec:localphase}

The information of the negative frequencies of the Fourier transform of a real-valued signal is redundant with respect to the positive frequencies. To remove the negative frequencies, D. Gabor proposed in 1946 the \textit{analytic signal} for which we refer to the excellent treatise \cite{smith-2007}. Using of analytic signal instead of the original real-valued signal has mitigated the estimation biases and eliminated cross-term artifacts due to the interaction of negative and positive frequencies~\cite{kay-1978}. However, the most interesting property of the analytic signal is its phase representation. In contrast to amplitude-based computer vision  techniques, the phase-based methods are not sensitive to smooth shading and lighting variations~\cite{boukerroui-noble-2004,granlund-knudsson-1995}. Moreover, beyond the global Fourier phase (not localized),  the analytic signal encodes simultaneously both local space and frequency characteristics of a signal. The phase-based feature detection has been investigated extensively in the classic computer vision approach, as in~\cite{boukerroui-noble-2004,moya-santacruz-2011, bayro-santacruz-moya-castillo-2016, moya-bayro-2014}. 
For a 1D real signal (function) $f(x)$, its analytical signal $f_{A}(x)$  is defined as follows 
\cite{granlund-knudsson-1995}:
\begin{equation}
    f_{A}(x)= f(x)-\ii f_{\mathcal{H}}(x)
\end{equation}
where $\ii = \sqrt{-1}$ and  $\mathcal{H}(f(x)) = f_{\mathcal{H}}(x)$ is the Hilbert transform of $f(x)$, namely
\begin{equation}
    f_{\mathcal{H}}(x)= \frac{1}{\pi} \int_{-\infty}^{\infty} \frac{f(\tau)}{\tau -x}d\tau.
\end{equation}
The amplitude $A$ and the phase $\varphi$ of $f_A$ are defined by the following expressions: 
\begin{eqnarray}
A(x) =&  \sqrt{f^2(x)+f_{\mathcal{H}}^2(x)}  \\
     \varphi(x) =& \arctan\left(\frac{f(x)}{f_{\mathcal{H}}(x)}\right).  
\end{eqnarray}
The local phase computation needs an additional operator to enhance  the localization of the features \cite{boukerroui-noble-2004}. One way to enhance the localization is to compute a filtered version of the signal $f(x)$ with an even function $f_{e}(x)$ as follows \cite{boukerroui-noble-2004}:
\begin{equation}
    f_{A}'(x)= f_{e}(x)*f(x)-\ii \mathcal{H}(f_{e}(x)*f(x)),
\end{equation}
where $*$ represents the convolution operator.
and $\mathcal{H}$ is the \textit{Hilbert transform}.
According to \cite{granlund-knudsson-1995,boukerroui-noble-2004} the approximation filter $f_e$ must be a band pass filter and  symmetric (even).  In practice, the approximation of the local phase and the local amplitude uses a pair of band-pass quadrature filters such as log-Gabor.

\subsection{Monogenic signal}

The most accepted 2D generalization of the analytic signal, is the  \textit{monogenic signal}. This notion was proposed by  Felsberg and Sommer in \cite{felsberg-sommer-2001}. According to them, monogenic signals are those that satisfy the generalized Cauchy-Riemann equations of Clifford analysis. 

A monogenic signal $I_{M}=I_{M}(x,y) \in \langle 1,\ii,\jj \rangle \subset \mathbf{H}$  associated to an image  $I=I(x,y) \in \mathbf{R}$ (where $x,y \in U$, $U$ a region of $\mathbf{R}^2$) is defined as follows \cite{felsberg-sommer-2001}:

\begin{equation}
    I_{M} = I' + I_{R},\quad I_{R} = \ii I_{1}+ \jj I_{2}, \label{eq:monogenic}
\end{equation}
where the signals $I_1$ and $I_2$ are the \textit{Riesz transforms} of $I$ in the $x$ and $y$ directions. In  this work, we apply a quadrature  filter approximation, by using  $I'= g*I$, where $*$ is the convolution operator and $g =g(x,y)$ is a \emph{log-Gabor} function (radial, isotropic,  bandpass filter).

Rewriting  equations of the monogenic signal (adding quadrature filters) in the Fourier domain we have:
\begin{equation}
I_{M} = \mathcal{F}^{-1}(J' + J_{R}),\quad J_{R} = \ii J_{1}+ \jj J_{2}, \label{eq:monogenicF}
\end{equation}
where $\mathcal{F}^{-1}$ is 2D inverse Fourier transform, $J=\mathcal{F}(I)$, $J'=J~\cdot~G$, $J_1 = J\cdot H_1 \cdot G$ , $J_2 = J \cdot H_2 \cdot G$, with:   

\begin{align}
& H_1(u_1,u_2) =\frac{u_1}{\sqrt{u_1^2+u_2^2}}, \\ \label{eq:RieszFouierd}  
& H_2(u_1,u_2) =\frac{u_2}{\sqrt{u_1^2+u_2^2}}, \\
& G(u_1,u_2) =   \exp\left(- \frac{ \log\left( \frac{\sqrt{u_1^2+u_2^2}}{\omega^s} \right)^2 }{2 \log({\sigma})^2}  \right), \\
& \omega^s = \frac{1}{\omega f^{s-1}} \label{eq:logGabor}
\end{align}
 Here $u_1,u_2$ are the frequency components, $\sigma$ is the variance of the log-Gabor,  $\omega$ is the minimum wavelength, $f$ is a scale factor, $s$  is the current scale.  
 
\begin{figure}[ht]
\centering
\includegraphics[width=0.6\textwidth]{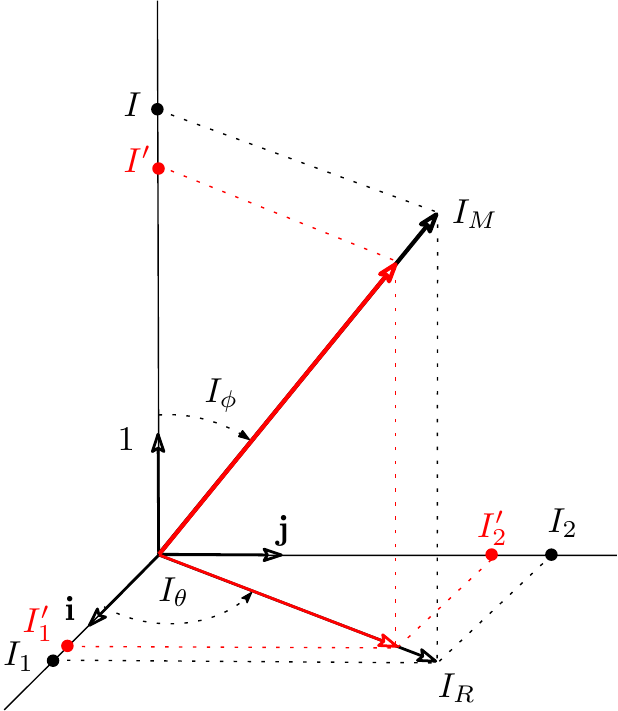}
\caption{Geometry of the monogenic signal. The changes in  $I'$ representing one pixel value intensity doesn't  affect the value of  $I_{\theta}$ and $I_{\phi}$.}
\label{fig:3D_mono}

\end{figure}

The \textit{local amplitude} $A_{M}=A_{M}(x,y)$ is defined (see \cite{felsberg-sommer-2001}) by the expression 
\begin{equation}
A_M={\sqrt{I'^2+ I_{1}^2+ I_{2}^2}}. \label{eq:mag_mono}
\end{equation}
The \textit{local phase} $I_{\phi}$ and the \textit{local orientation}  $I_{\theta}$  associated to $I'$  are defined, again following \cite{felsberg-sommer-2001}, by the relations 
\bea
I_{\phi} &=& \atantwo \left( \frac{I'}{|I_R|}\right),  \label{eq:Ipsi}  \\
I_{\theta} &=& \atan{\left(\frac{-I_2}{I_1}\right)}, \label{eq:Itheta} 
\eea
where $|I_R| = \sqrt{I_1^2+ I_2^2}$ and the quotients of signals are taken point-wise.
The geometric interpretation of the monogenic signal is depicted in  Figure \ref{fig:3D_mono}. Note how the changes in the pixel value intensity $I'$ doesn't  affect the value of the local phase $I_{\theta}$ and the local orientation $I_{\phi}$. This theoretically invariant response to large illumination changes of the local phase is in line with the reported at \cite{boukerroui-noble-2004,moya-xambo-perez-salazar-mzortega-cortes-2020}

\subsection{Bio-inspired properties and tools}

In this subsection we highlight the main properties of the V1 cortex layer and how are they reflected in the functionality of M6. We also outline the reasons for choosing two bio-inspired technical tools as key ingredients for guiding us in its design and construction.

The V1 cells form the first layer of the hierarchical cortical processing~\cite{bigun-2006}. The analogy for M6, quite literal, is that it is meant to be the first layer of a ConvNet. 

The V1 neurons respond vigorously only to edges (odd-signal) and lines (even-signal) at a particular spatial direction through their orientation columns~\cite{granlund-knudsson-1995}. The counterpart of this in M6 is realized by computing the local phase and orientation, as these have the capability of detecting lines, edges and their orientations. This also justifies the placing of M6 as the first layer, as the features in question are the most primitive and hence it would be the wrong job for a layer in a deeper position.

In \cite{xing-yeh-2014}, the authors reported a cortical adaptation to brightness and darkness in the macaque primary visual cortex V1. In relation to this, it is worth mentioning that the experimental results of the present work confirm the expectation advanced in  \cite{granlund-knudsson-1995,boukerroui-noble-2004, moya-xambo-salazar-sanchez-cortes-2021} that the local phase and local orientation should be invariant with respect to brightness and contrast changes due to its geometry representation. 

Daugman \cite{daugman-1985} discovered that the Gabor functions  resembled the experimental findings of Hubel and Wiesel \cite{hubel-wiesel-1962} on orientation selectivity of the visual cortical neurons of cats. However,  Gabor functions (filters) may cause fairly poor pattern retrieval accuracy in certain applications (see \cite{nava-escalante-cristobal-2012}) because it has, for certain bandwidths, an undesirable non-zero value of the so-called DC component (cf. \cite{nava-escalante-cristobal-2012}). For pattern recognition applications, this DC component entails that a feature can change with the average value of the signal. Fortunately, this weakness can be overcome with log-Gabor filters, as explained in \cite{nava-escalante-cristobal-2012}, and this is why we use them henceforth.





The other bio-inspired tool is the HSV color space. Although the main virtue of this space is that it best fits the human perception of color \cite{smith-1978}, in this work we use it as a means to geometrically code the phase and orientation in the Hue channel. This is quite natural, as the purpose of this channel is to hold a phase. The use of this transformation in the M6 layer results in an increase of classification accuracy (see Figure \ref{fig:m6comp}).


\section{Monogenic layer  M6}\label{sec:mono}

{The  monogenic signal  characteristics (contrast and illumination invariance based on its geometry, feature extraction in the space and frequency domains, and its similarities to the V1 properties) are the main stimuli for the design of our M6 trainable (top) layer.}  

\begin{figure}[ht]
\vspace{-9pt}
\centering
\includegraphics[width=0.95\textwidth]{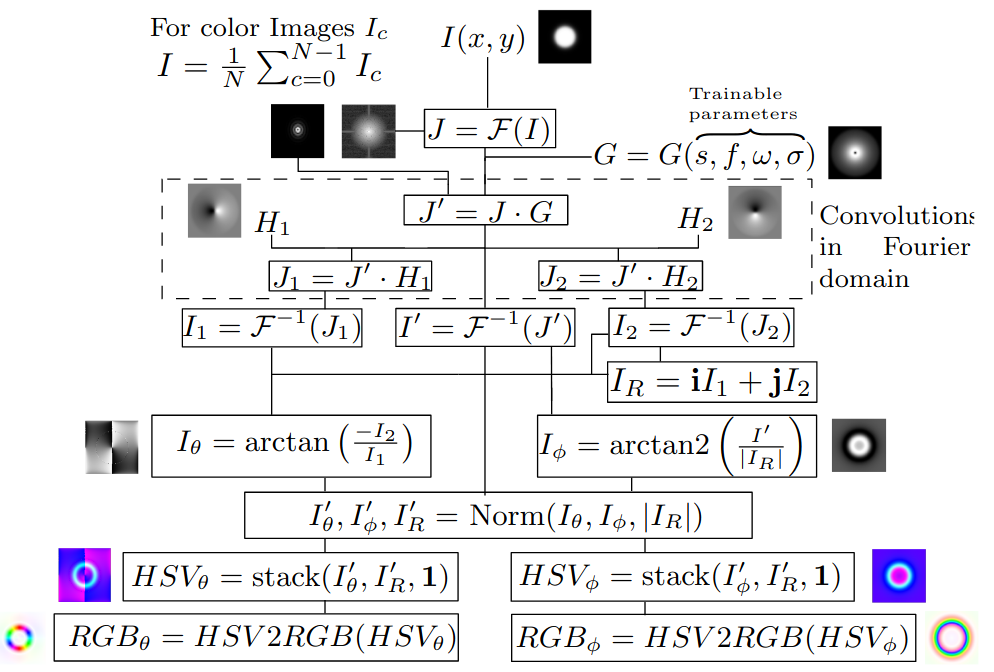}
\caption{Computational flow of an M6 unit layer and examples (images) of the computation outputs. See the text below for details.}
\label{fig:m6layer}
\end{figure}

A scheme of the computational flow of the M6 unit on a one-channel image as input is displayed in Figure \ref{fig:m6layer}. For color images ($I_c$) we compute the mean value of the channels and use it as input value. The convolution operations are done in the Fourier domain, that is, on $\mathcal{F}(I)$, where $I$ is the input image. 
The reasons for implementing the convolutions in the Fourier domain are that it allows a straightforward implementation of the monogenic signal and avoiding the problem of having to select a convolution kernel size as we can proceed with a band-pass filter instead.

It is fundamental to note that M6 only has four trainable parameters $(s, f,\omega, \sigma)$, which are the parameters of the log-Gabor function (eq. \eqref{eq:logGabor}). Recall that $s$ is the current scale indicator; $\sigma$, the variance of the log-Gabor; $\omega$, the minimum wavelength; and $f$ a scaling factor.  

\begin{figure}[ht]
\vspace{-9pt}
\centering
\includegraphics[width=0.89\textwidth]{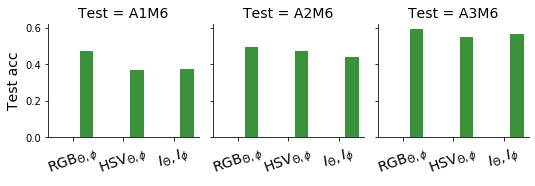}
\vspace{-9pt}
\caption{Test accuracy values over the three models with CIFAR-10 using different elements of M6: RGB output, HSV output and the phases.}
\label{fig:m6comp}
\end{figure}

Although the local phase $I_{\phi}(x,y)$ (Eq. \ref{eq:Ipsi}) and local orientation $I_{\theta}(x,y)$ (Eq. \ref{eq:Itheta}) are theoretically invariant to illumination changes, it turns out that to better mimic the V1 behavior, and to increase the classification performance, it is beneficial to insert additional processing operations, as explained below. In Figure \ref{fig:m6comp} we present an example showing how the performance increases by leveraging the RGB phases.

After the local phase computation, we add a normalization step, $I'_{\theta},I'_{\phi},I'_R=\text{Norm}(I_{\theta},I_{\phi},I_R)$, where Norm is generically defined by the expression
\begin{equation}
\text{Norm}(I) = \frac{I(x,y)-\min(I(x,y))}{\max(I(x,y))-\min(I(x,y))}. \label{eq:max-min-norm}
\end{equation}
Next, using the polar geometry of the HSV color space, we stack the normalized local phase $I'_{\phi}$ and local orientation $I'_{\theta}$ in a hue channel (H) of a HSV color space with  $I'_{R}$ as saturation (S) and the constant matrix $\mathbf{1}_{[m,n]}$ as the value component (V). Finally, the HSV images are converted by the standard function $HSV2RGB$ (see \cite[page 304]{agoston-2005}) into RGB images.

\begin{figure}[ht]
\vspace{-9pt}
\centering
\includegraphics[width=0.8\textwidth]{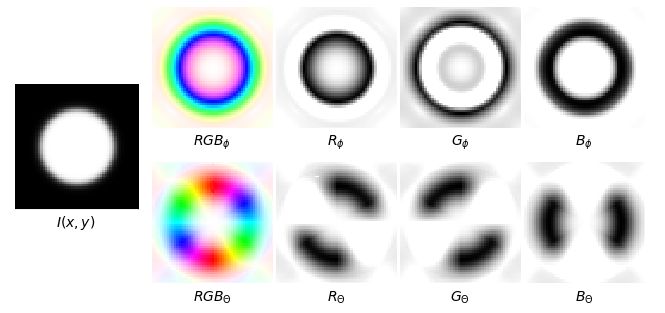}

\vspace{-9pt}
\caption{The white circle is the input image ($I(x,y)$) and $RGB_{\phi}$ and $RGB_{\theta}$ are the M6 outputs. The grey images are the corresponding RGB components.}
\label{fig:m6outs}
\end{figure}

\begin{figure}[ht]
\centering
\includegraphics[width=0.75\textwidth]{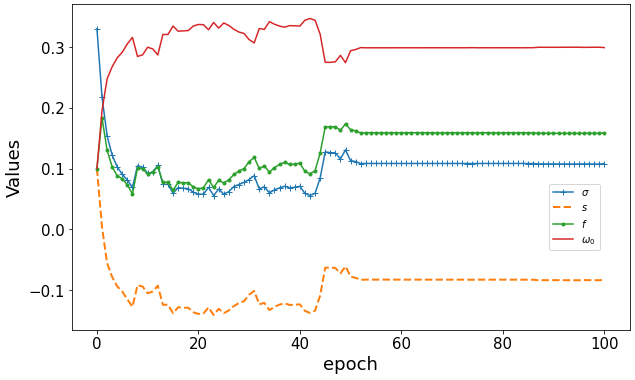}
\caption{Example of the evolution of the M6 trainable parameters ($s,f,\omega,\sigma$) on the MNIST dataset.} 
\label{fig:parameters_one}
\end{figure}

\begin{figure}[ht]
\centering
\includegraphics[width=0.75\textwidth]{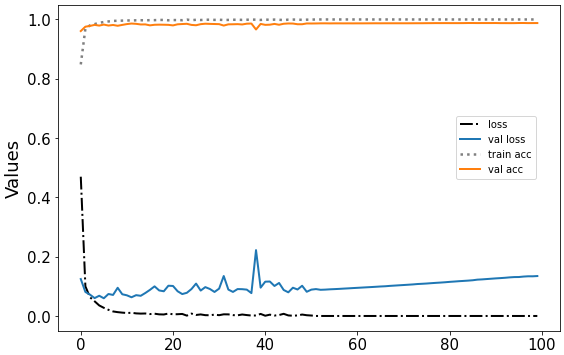}
\caption{Example of the evolution of the loss, validation loss, accuracy and validation accuracy on the MNIST dataset and the M6 layer.} 
\label{fig:parameters_zero}
\end{figure}

\begin{figure}[ht]
\centering
\includegraphics[width=0.78\textwidth]{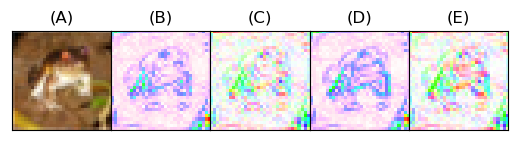}
\caption{Example of the activations of the M6 unit before and after training. (A) Input image; (B) and (C), activations $RGB_{\phi}$ and $RGB_{\theta}$ before training; (D) and (E), the same activations after training. Note that (D) and (E) are shaper than (B) and (C), respectivetly.} 
\label{fig:act_tr}
\end{figure}

As a result, an M6 unit produces six output feature maps per one channel image input. Figure \ref{fig:m6outs} depicts an example of M6 outputs using a white circle as input image. 
About the backpropagation, we used the automatic differentiation and weight updating implemented by Tensorflow 2 (TF), but their symbolic expression is not reproduced here because it lies beyond the scope of this paper. Instead, we provide two graphics that document this. Figure \ref{fig:parameters_one} illustrates how the M6 trainable parameters are adjusted during the training process using the MNIST dataset. We note that the major changes in the parameters take place in the first fifty epochs. Figure \ref{fig:parameters_zero} displays the accuracy and loss in the training  and validation processes corresponding to the same training job. The validation loss (val loss) and validation accuracy (val acc) undergo the major changes before the fiftieth epoch, thus matching what was found in Figure \ref{fig:parameters_one}. Notice that the validation loss rises slightly after the fiftieth epoch,
a sign that it has entered the overfitting regime. In addition, Figure \ref{fig:act_tr} shows an example from another perspective, namely a CIFAR-10 image and the associated feature maps before and after training, which reveal that the feature maps after training are sharper. This behavior is expected inasmuch as the trainable parameters define the band pass size of the log-Gabor filter. 

\subsection{M6 properties and comparison with a regular convolution layer}

Table \ref{tab:CNNL_comparison} summarizes  a comparison of some characteristics of a convolution layer with those of the proposed layer $M6$. One of the  main differences between a first layer C of a conventional ConvNet and the proposed layer M6 is that the convolutions in the latter are carried on in the Fourier domain. Furthermore, the number of trainable parameters of M6 (four) is significantly lower than the number of parameters of C, which depends on the number and size of its filters. For example, for six $3\times3$ filters, the number of parameters is $6\times 3^2=54$.  On the other hand, if both C and M6 can detect lines (l) and edges (e), usually C can detect more features, such as corners (c) or some irregular shapes, whereas M6 is crucially also sensitive to orientations (thus resembling V1). The orientation response is important to obtain a robust performance even with rotated images. To note also an important  difference in the activation functions, with ReLU (Rectifying Linear Unit) having become very popular in C instances \cite{goodfellow-bengio-courville-2016}, while in M6 that role is played by $\arctan$ and $\arcsin$. This notwithstanding, both M6 and C behave similarly with respect to all the tested optimizers (for instance, SGD, ADAM and NADAM).
A final fundamental remark is that the trainability of M6 sets it apart from deterministic (pre-processing) layers used in other systems, as stressed in Section \ref{sec:related}. 

\begin{table}[ht]
\begin{center}
\caption{Comparison of  a regular convolutional layer C, and the M6 layer.\label{tab:CNNL_comparison} }
\begin{tabular}{lcc}
\toprule
Characteristics &  C   &   M6   \\
\midrule
Convolution      & Space & Frequency  \\
Parameters      & (3@3)@6=54 & 4  \\
Feature elements & l, e, c, etc & l and e  \\

Oriented response & No  &  Yes     \\
Nonlinear function & ReLU &  $\arctan$, $\arcsin$  \\
Layer position & Any  & First   \\
Trainable &  Yes & Yes  \\
\bottomrule
\end{tabular}
\end{center}
\end{table}

\section{Experimental Setup} \label{sec:exp_setup}

The experimental setup is designed to evaluate the robustness and the classification performance under different datasets, architectures and degradations.


\subsection{Datasets}\label{subsec:datasets}
We use four datasets: MNIST \cite{lecun-cortes-burges-2010}, Fashion-MNIST (f-MNIST) \cite{han-kashif-2017}, CIFAR-10   \cite{krizhevsky-Hinton-etothers-2009}, and Dogs vs Cats (DvsC) \cite{elson-doucer-2007}. All datasets are available online through the Tensorflow datasets (tfds)~\cite{abadi-et39-2016}  package. The haze degradation dataset is available through the Gitlab link provided below.  Table \ref{tab:data}  shows how we split the datasets and their main characteristics.


\begin{table}[ht]
\begin{center}
\caption{Split size and input shape of the datasets} \label{tab:data}
\begin{tabular}{lcccc}
\toprule
 & MNIST & f-MNIST & CIFAR-10 & DvsC \\
\midrule
Training set   & 48,000 & 48,000 & 40,000 & 16,284 \\
Validation set & 12,000 & 12,000 & 10,000 & 3,489 \\
Test set & 10,000 & 10,000 & 10,000 & 3,489 \\
Total & 70,000 & 70,000 & 60,000 & 23,262 \\
Input shape & [28x28x1] & [28x28x1] & [32x32x3] & [128x128x3] \\
\bottomrule
\end{tabular}
\end{center}
\end{table}

\subsection{Degrading procedures} 
\begin{figure}[ht]
\vspace{-12pt}
\centering
\includegraphics[width=0.98\textwidth]{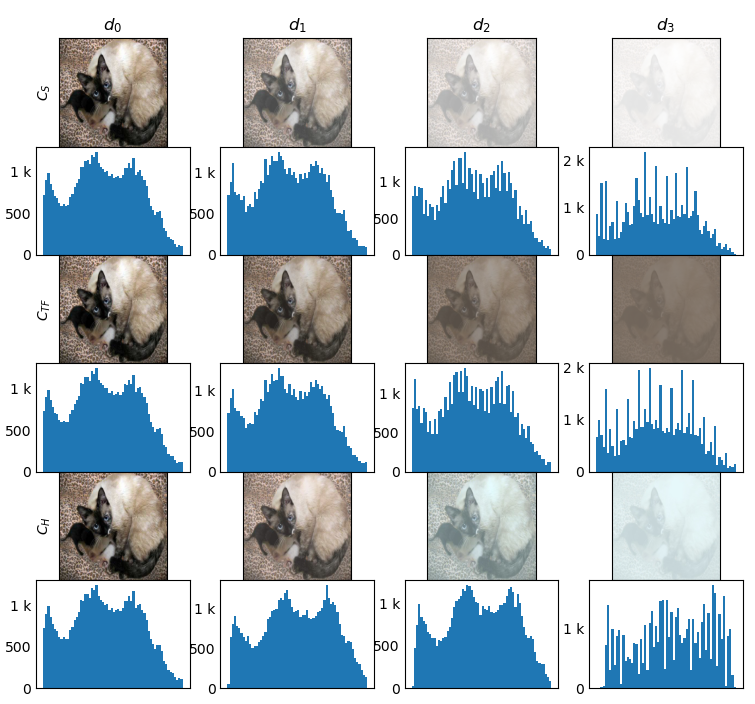}
\caption{Examples of the three contrast degradation procedures $C_{S},C_{TF},C_{H}$ applied to the original image (column $d_0$, no degradation) for the degrading levels $d_j$ ($j=1,2,3$) described in the text.}
\label{fig:deg_hist}
\end{figure}

We use three contrast transformations to degrade images $I$.  The max-min scale transformation, $C_{S}I$;  The TF contrast transformation, $C_{TF}I$; and the haze transformation, $C_{H}I$. Figure \ref{fig:deg_hist} displays an example of three degradation levels $d_j$ ($j=1,2,3$) for each of the degradation procedures applied to the image in column $d_0$ (no degradation). 

\smallskip\noindent
\emph{Max-min scale transformation.}
Applied to an image $I(x,y)$ with respect to an interval $S=[a,b]$, it produces a image  
\begin{equation}
    C_{S}I(x,y) = a + \frac{(I(x,y) - \text{min}(I(x,y))(b-a)}{\text{max}(I(x,y))-\text{min}(I(x,y))}.
    \label{eq:max_min_scale}
\end{equation}
In our experiments we use this transformations with the four degrading levels $d_j$  corresponding to the intervals $S$ defined by, respectively, $a= 0.3, 0.7, 0.9$ and $b= 1$. 


\smallskip\noindent
\emph{Contrast degradation using $C_{TF}$.}
It is defined by  
\begin{equation}
    C_{TF}I(x,y) = \mu + F(I(x,y)- \mu), \label{eq:Ctf}
\end{equation}
where $\mu$ is the mean value of the input image $I$ and $F\in [0,1]$ is a contrast factor. In our experiments, we take four degradation levels $d_j$ corresponding, respectively, to the values $F\in\{0.7,0.3,0.1\}$.

\smallskip\noindent
\emph{Contrast degradation by haze.}
The idea of this method of contrast degradation is adding non-uniform haze to a given image. The main basis for gauging haze and fog in images stems from the atmospheric scattering model proposed by \cite{McCartney-Hall-Freeman-1977} and which can be summarized, following \cite{he-sun-tang-2011}, by the equation
\begin{equation}
C_{H}I(x,y) =  t(x,y) I(x,y)+(1-t(x,y))A(x,y). \label{eq:atms}
\end{equation}
In this expression $I(x,y)$ stands for what the image would be if haze were removed and $C_{H}I(x,y)$ is the observed hazed image. The rationale behind \eqref{eq:atms} is that $I(x,y)$ undergoes an attenuation $t(x,y) I(x,y)$ caused by the \textit{medium transmission} $t(x,y)\in[0,1]$, which measures the fraction of light reaching the camera from the $(x,y)$ direction. Its value is 1 for a perfectly transparent air and 0 when no light from $(x,y)$ reaches the camera. The term $A(x,y)$ stands for the \textit{total atmospheric light}, and hence $(1-t(x,y))A(x,y)$ measures the fraction of light contributing to $C_{H}I(x,y)$ not originated at the source $(x,y)$, usually produced by scattering and reflection processes.

Now the main point in our contrast degrading is using equation \eqref{eq:atms} the other way around, which means generating various  values of $A$, use these and $I(x,y)$ to estimate $t(x,y)$ by the dark channel prior proposed in \cite{he-sun-tang-2011}. In our experiments, the degrading levels $d_j$  are defined by choosing, respectively, $S=[a=0.5, b=0.8], [a=0.3, b=0.5], [a=0, b=0.15]$. 


It remains to see how do we generate the various $A$.
The components of an RGB image $I$ ($I^r$, $I^g$, $I^b$) allow us to regard $I(x,y)$ as a 3-vector, 
$I(x,y)=[I^r(x,y),I^g(x,y),I^b(x,y)]$. In this case, $A(x,y)$ has to be also a 3-vector, say 
$A(x,y)=[A^r(x,y),A^g(x,y),A^b(x,y)]$, and as a result $C_HI(x,y)$ can be treated in the same way. Our generation of the $A$ vectors is done by choosing each of its channels $c$ independently at random in the interval $[0.8,1]$, $A^c\in[0.8,1]$. It is to be remarked that the contribution of $A$ is not limited to the light intensity, as it produces changes in color, a fact that is in accord with the physical effects of the atmospheric light.

\comment{
\begin{figure}[ht]
\centering
\includegraphics[width=0.46\textwidth]{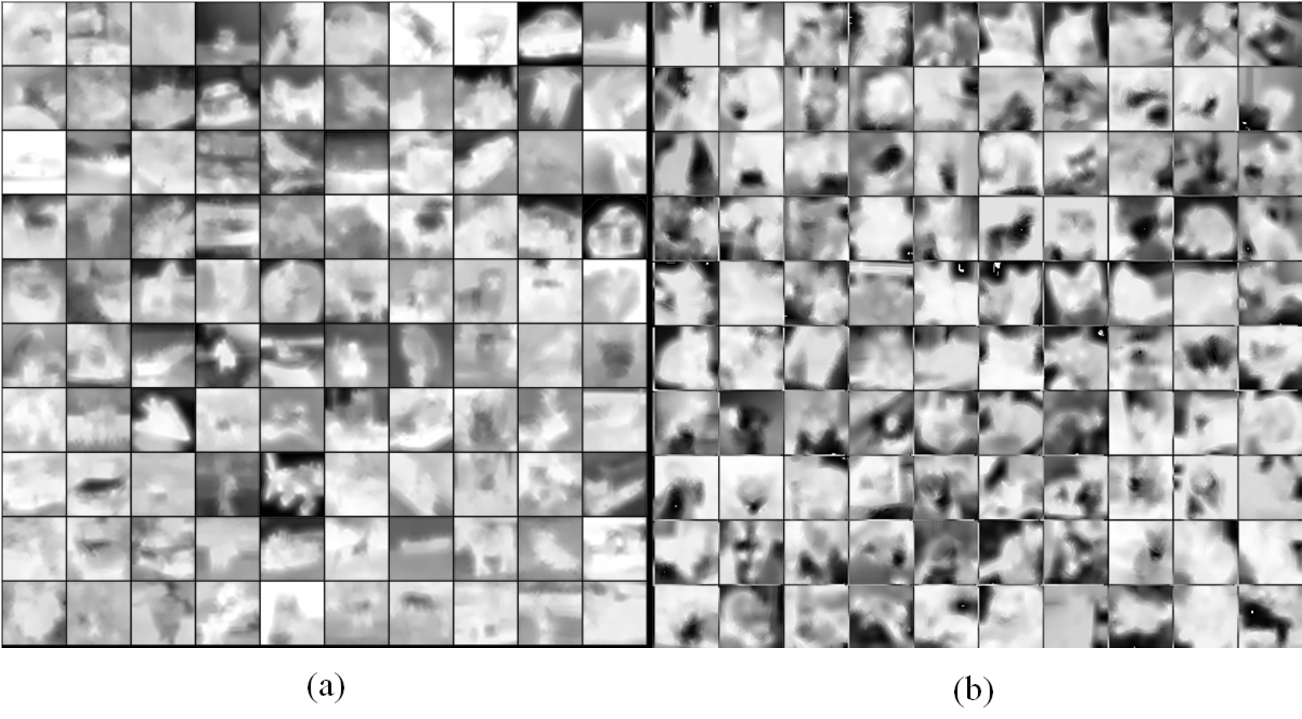}
\caption{Examples of transmission map estimation $t(x,y)$. (a) Transmission maps of 100 images from CIFAR-10; (b)  Transmission maps of 100 images from DvsC dataset.}
\label{fig:dc}
\end{figure}

\begin{table}[ht]
\begin{center}
\caption{Degradation parameters of color of atmospheric light $A(r,g,b)$ and  the transmission map $t(x,y)$.} \label{tab:degradation}
\begin{tabular}{ cl }
\toprule
\multicolumn{2}{ c}{$A([0.8,1],[0.8,1],[0.8,1])$}  \\
\midrule
Level & Parameters  \\
\midrule
$d_0$ &  zero degradation  \\
$d_1$ & $t(x,y)=[0.5,0.8]$  \\
$d_2$ & $t(x,y)=[0.3,0.5]$   \\
$d_3$ & $t(x,y)=[0.0,0.15]$   \\
\bottomrule
\end{tabular}

\end{center}

\end{table}
}

\subsection{Architectures, training and test}
\begin{figure}[ht]
  \centering
  \includegraphics[width=0.98\textwidth]{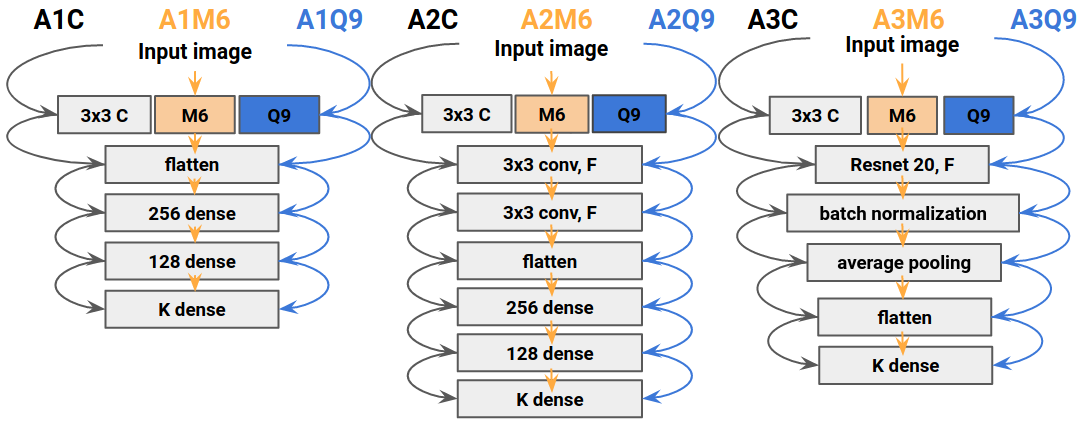}
  \caption{Triple architectures (AjC, AjM6, AQ9) used for training, validation and test. The input image shapes are described in Table \ref{tab:data}. The terminology used follows the conventions of the TF framework. For more details, see the source code available through the link provided below. }\label{fig:arch}
\end{figure}

The three architectures used in this work are labeled A1, A2 and A3. A1 is a shallow ConvNet; A2, a medium depth ConvNet; and A3, a deep ConvNet that grafts a ResNet20 \cite{he-mcauley-2016} as a subnet or backbone. Each of these architectures actually stands for tree: one, AjC, topped by a standard convolution layer (C),  other, AjQ9, topped by a Q9 layer defined at\cite{moya-xambo-perez-salazar-mzortega-cortes-2020}  and AjM6, topped by the M6 layer. See Figure \ref{fig:arch} for details about each of them. It is  fundamental to note that we compare the classification robustness of the M6 layer against the conventional ConvNet layer (kernel $3\times3$, with six output channels, and no data augmentation in the training process). The $K$ in the last layer (with a softmax function), means the  size of the output layer  (number of classes), which is 10 in for CIFAR-10, MNIST, and f-MNIST and 2 for DvsC.  It is important to remark that the large depth of A3  does not allow to process small size images such as those of MNIST and f-MNIST. 




To  evaluate  the response robustness,  the  models  were  trained with  a  specific  data  degradation  and  tested  with not only the same data degradation but also with  three additional data degradations. This kind of experimental setup is common  in other works to evaluate robustness, such as \cite{moya-xambo-perez-salazar-mzortega-cortes-2020, hendrycks-dietterich-2019,dapello-marques-schrimpf-geiger-cox-dicarlo-2020}.
The experimental setup concerns the four primary datasets mentioned before, the nine networks summarized in Figure \ref{fig:arch}, and the three methods of contrast degradation. For each method of degradation and each primary dataset, we construct three additional datasets by applying the degradation method to the primary dataset with degradation levels $d_1$, $d_2$ and $d_3$. Now, given one of the networks, train it four times, one for each of the degraded datasets, including the primary dataset (degradation level $d_0$). Note that all the training was made from scratch. Finally, test each of the four trained classifiers on the four degraded datasets. 
These arrangements are sketched in Table \ref{tab:experiments}, where Tr stands for training (one for each degradation level). As indicated before, the $A3$-C and $A3$-M6 networks can be used only on CIFAR-10 and DvsC data.

\begin{table}[ht]
\begin{center}
\caption{Sketch of the experimental setup. } \label{tab:experiments}
\begin{tabular}{ cccccc }
\toprule
\multicolumn{2}{ c}{\textbf{$C_{TF}$}}& \multicolumn{2}{ c}{\textbf{$C_{S}$}} & \multicolumn{2}{ c}{\textbf{$C_{H}$}}  \\
\midrule
Tr & Test & Tr & Test & Tr & Test  \\
\midrule
$d_0$ & $d_0$, $d_1$, $d_2$, $d_3$ & $d_0$ & $d_0$, $d_1$, $d_2$, $d_3$ & $d_0$ & $d_0$, $d_1$, $d_2$, $d_3$      \\
$d_1$ & $d_0$, $d_1$, $d_2$, $d_3$ & $d_1$ & $d_0$, $d_1$, $d_2$, $d_3$ & $d_1$ & $d_0$, $d_1$, $d_2$, $d_3$   \\
$d_2$ & $d_0$, $d_1$, $d_2$, $d_3$ & $d_2$ & $d_0$, $d_1$, $d_2$, $d_3$ & $d_2$ & $d_0$, $d_1$, $d_2$, $d_3$  \\
$d_3$ & $d_0$, $d_1$, $d_2$, $d_3$ & $d_3$ & $d_0$, $d_1$, $d_2$, $d_3$ & $d_3$ & $d_0$, $d_1$, $d_2$, $d_3$ \\
\bottomrule
\end{tabular}

\end{center}
\medskip

\end{table}

The hyperparameters used in the experimental setup  are a learning rate of 0.001, 100 epochs, batch-size 128, ReLU as activation function for C, and Adam as optimizer.  It should be noted that we use Keras-turner (Bayesian optimization) \cite{chollet-et-al-2015} to choose the  kernel size and learning rate of the convolution layer. The initial parameters of M6 layer are $s=1$, $f=1$, $\sigma=0.33$, $\omega=1$ based on the previous version of the layer \cite{moya-xambo-salazar-sanchez-cortes-2021}. All datasets were normalized to one by dividing each pixel value by 255.  
We have relied on the TF 2.1 deep learning framework run on a V-100 Nvidia-GPU for all experiments.  Its reproducibility is supported by  the supplemental material uploaded at the Gitlab link  \href{https://gitlab.com/monogenic-layer-m6/monogenic-layer-trainablebio-inspired-cnnlayerforcontrastinvariance}{\underline{M6 project}}.




\subsection{Metrics for analysis}

An Structural Similarity Index Measure (SSIM) index was chosen in order to compare the effect of the degradation procedures on the M6 feature maps (in Section \ref{sec:results}). This index was selected because it allowed us to compare numerically the feature maps changes made by the degradations  taking into account some key aspects of human perception \cite{gao-rehman-wang-2011}. Moreover, SSIM  could quantify the image quality as the \textit{perceived} changes in the structural information (SSIM \textit{map}), and at the same time,  SSIM takes into account the  luminance  and contrast changes. The SSIM index formula is defined as follows \cite{gao-rehman-wang-2011} :
\begin{equation}
\hbox{SSIM}(x_1,y_1) = \frac{(2\mu_{x_1}\mu_{y_1} + c_1)(2\sigma_{x_1y_1} + c_2)}{(\mu_{x_1}^2 + \mu_{y_1}^2 + c_1)(\sigma_{x_1}^2 + \sigma_{y_1}^2 + c_2)},    
\end{equation}
where  $x_1$ and $y_1$ are two arrays of size $N\times N$, $\mu_{x_1}$ and $\mu_{y_1}$ are the averages, and  $\sigma_{x_1}^2$ and $\sigma_{y_1}^2$ the variances, of $x_1$ and $y_1$ respectively, while $\sigma_{x_{1}y_{1}}$ is the  covariance of $x_1$ and $y_1$, $c_1=(k_1L)^2 $ and $c_2=(k_2L)^2$, with $L$ the value of the dynamic range, $k_1=0.01$ and $k_2=0.03$. The value of SSIM belongs to the interval $[0,1]$, with SSIM$=1$ corresponding to maximum similarity and SSIM$=0$ to minimum similarity.

\section{Results and Analysis}\label{sec:results}
\begin{figure}[ht]
  \centering
  \includegraphics[width=0.89\textwidth]{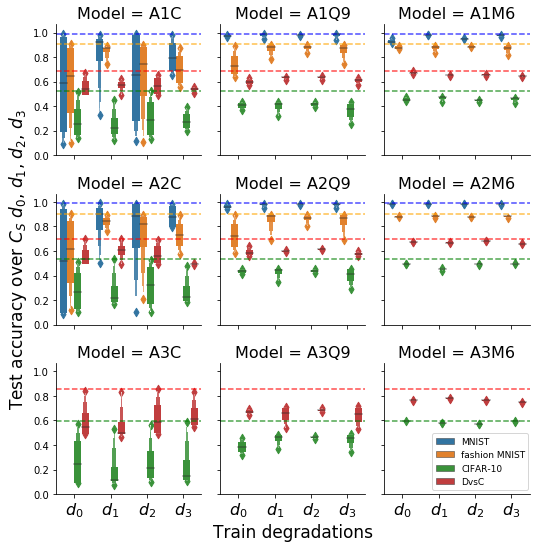}
  \caption{Boxen plots of test accuracy over $C_{S}$ degradations for all model and datasets combination. See the text for more details.}\label{fig:Cs_results}
\end{figure}

\begin{figure}[ht]
  \centering
  \includegraphics[width=0.89\textwidth]{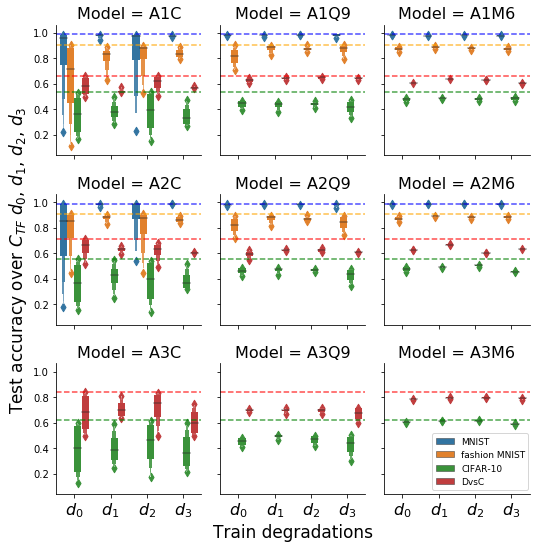}
  \caption{Boxen plots of test accuracy over $C_{TF}$ degradations for all model and datasets combination}\label{fig:Ctf_results}
\end{figure}

\begin{figure}[ht]
  \centering
  \includegraphics[width=0.89\textwidth]{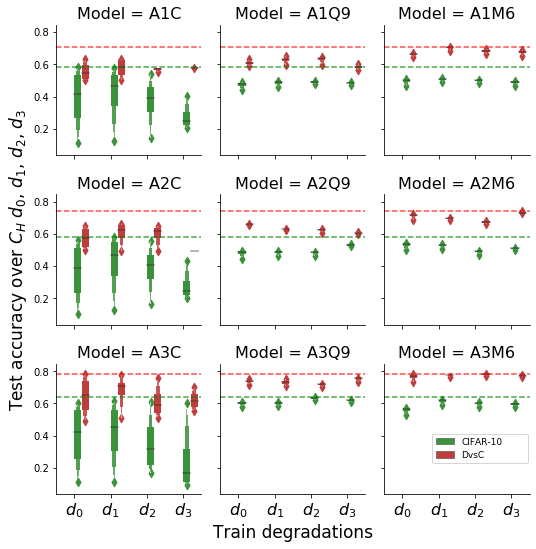}
  \caption{Boxen plot of test accuracy over $C_{H}$ degradation for all model and datasets combination}\label{fig:Ch_results}
\end{figure}

In Figures \ref{fig:Cs_results}, \ref{fig:Ctf_results}, and \ref{fig:Ch_results}  we present a grid of boxen plots of the test-accuracy values of the classification according to our experimental setup.  We compare C and Q9 against M6 using four datasets (color), three architectures (columns) and four levels of degradation (using $C_{S}$, $C_{TF}$, and $C_H$, respectively). Each box represents four test accuracy values associated with $d_0,d_1,d_2,d_3$. For instance, the  A1C model, trained with $d_0$ for MNIST, has four test accuracy values using $C_{S}$:  $d_{0}=0.986$, $d_{1}=0.953$, $d_{2}=0.225$, $d_{3}=0.089$.  Their box plot reflects the computed (quartile) values  $Q_0 = 0.089$,  $Q_1 = 0.191$, $Q_2 = 0.590$,  $Q_3 = 0.961$ and  $Q_{4} =0.986$.  The dash lines represent the maximum test accuracy value. The missing boxes for MNIST and f-MNIST with $A3$ are due to the fact that their input sizes are incompatible with the large depth of that architecture.  

On a wider angle, it is possible to divide the analysis of these results in two major classes: i)  evaluating the robustness  using the size of the box and whiskers, and ii) evaluating the maximum performance with the top whisker.  For i), these tests  highlight that the ConvNets with the M6 present the smallest boxes for all the cases. These results provide indisputable evidence on the robustness effect of M6 under different contrast degradations. Please note that we train on each degradation level and test with all degradation levels and we find that M6 can be trained with any degraded level and yet achieving almost the same performance on the others. This is in marked contrast to the C models, as they have quite low performance when trained with $d_3$ degraded images.
To facilitate the view of the maximum values, ii), we add dash lines with the maximum value for each model type (A1, A2 and A3) and each dataset. The maximum performance comes about 17/26 for C,  3/26 for Q9 and  6/26 for the M6  model. However, with a few exceptions, the ConvNets with M6 have a performance closer to the maximum test accuracy value. Computing the square of the difference with respect to each maximum, we get 29.3, 4.7, and 1.0 for the C, Q9 and M6 architectures, respectively. It is important to note that the maximum values from the C models are generally achieved when the training uses the same data degradation  as the test dataset. However, for the maximum degradation level $d_3$, the classification performance is significantly lower even with the same data degradation level as that used for training.

The results also reveal parallel behavior  patterns between the  C and M6 models. For instance, all models  have higher test accuracy performance with simple datasets (MNIST and f-MNIST). In addition, the performance tends to increase with the depth of the architecture. However, it is important to remark that the single M6 unit, which has just four weights, can extract enough features for different types of datasets in contrast to the C layer with six convolution units  and fifty-four weights. 
In addition,  the increase
in the test accuracy of M6 in relation to the increase of the number of layers can be ascribed to the fact that the features output by M6 can be learned and combined by the later layers much like the way they are processed by a conventional convolutional layer. Combining the robustness and the maximum classification performance of M6 leads to boosted efficacy of those layers 
when fed with the M6 output rather than having to elicit it from a raw input. In other words, M6 delivers sharper features for image classification.

\begin{figure}[ht]
\centering
\includegraphics[width=0.695\textwidth]{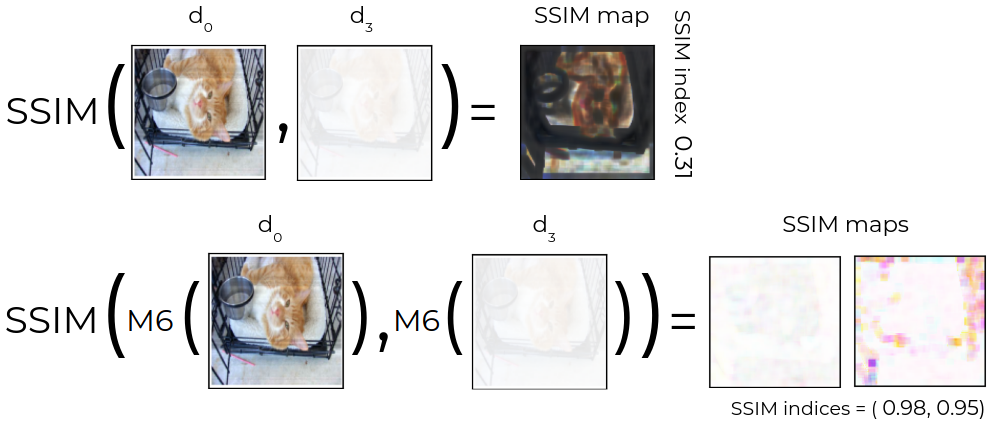}
\caption{Example of SSIM map and index when applied to $d_0$ and and one $d_3$ degradation of it (top row) and the SSIM maps and indexes of M3($d_0$) and M3($d_3$). See the text above for the interpretation of these data.}
\label{fig:ssim_ex}
\end{figure}

As stressed in the description of M6, the robust performance stems from the geometry behind its design. One way to evaluate the resilience to contrast changes is using the SSIM index. The main idea is to compute the SSIM index for $d_0$ and $d_j$ images ($j=1,2,3$) and compare it with the two SSIM indexes of the image transformed by M6 (which we regard as two images, $RGB_{\theta}$ and $RGB_{\phi}$). The comparison is even more forceful if the indexes are displayed together with the SSIM maps. In the top row of of Figure~\ref{fig:ssim_ex}, for example, SSIM is applied to a $d_0$ image and its $d_3$ degradation by $C_S$, and it is not surprising to find a low index, 0.31, and a discrepancy map that is quite close to the original image. In the bottom row, SSIM is applied to the same images after being transformed by M6, and we see that the two indexes are high (close to 1) and that the discrepancy maps have very small values, which means that M6 drastically reduces the discrepancy between M6($d_0$) and M6($d3$). This behavior is a compelling evidence of the capability of M6 and suggests the potential of layers designed on similar, possibly more general principles. 

\begin{figure}[ht]
\centering
\includegraphics[width=0.69\textwidth]{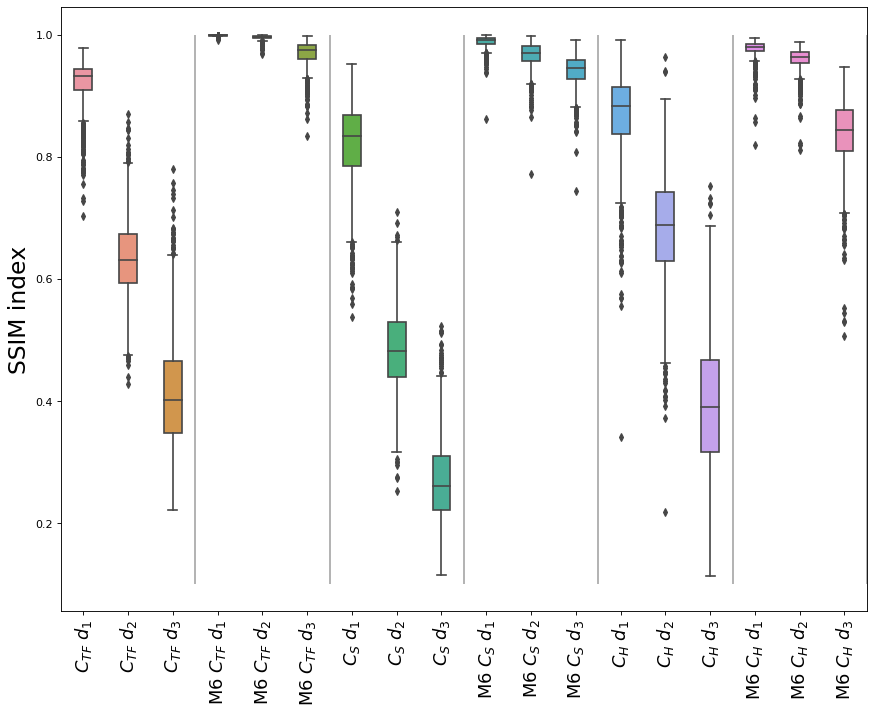}

\caption{SSIM index of a thousand $d_0$ images with their degradations $d_j$ ($j=1,2,3$) using $C_{S}$, $C_{TF}$, $C_{H}$,
together with similar computations after being transformed by M6.}
\label{fig:ssim}
\end{figure}

Figure \ref{fig:ssim} shows a box plot of the results of a similar computation, but this time involving 1000 random images,
three degradation methods ($C_{S},C_{TF},C_{H}$) and three degradation levels $d_j$ ($j=1,2,3$). It is clear that the box plots of the values for the M6 transformations are more compact and quite close to one. This confirms what we found in the preceding example, namely that 
M6 tends to see the $d_j$ degraded images as a $d_0$ image.


In order to compare the performance analysis of M6  against C, we  average the time  and memory consumption in the  training  and the testing. The results are that M6 achitectures spend 7\% more time in the training and 26\%  more time in testing. The memory consumption of M6 turns out to be 5 times higher. We impute this difference to the fast Fourier transform  computation. To optimize this memory consumption further work will be needed. To remark also that we cannot compare meaningfully with the Q9 performance because its code is only available for CPU while the M6 runs on GPU.

\section{Conclusions and Future work} \label{sec:conl}

We have designed and presented a new trainable bio-inspired front-end layer for ConvNets. This new layer generates a 3D geometric representation of each pixel value by computing the quaternion monogenic signal in the Fourier domain. As a result, it is possible to leverage the local phase and the local orientation to elicit low-level geometric features, such as oriented lines or edges. The coupling of the proposed layer with a regular ConvNet, or with dense networks, achieves a classification of images that is little affected by severe contrast degradings and which, on the whole, has better accuracy. The experimental results are consistent with the geometrical observation that the local phase and the local orientation are invariant to variable contrast conditions.


Concerning the impact of our work, it is to be searched in situations where an invariant response to contrast alterations is required. Among the possible scenarios we count self-driving cars under haze conditions, surface glazes in medical images (biopsies), or day-round autonomous video surveillance.  



We hope that our research will serve as one of the lines for future studies on equivariant and invariant representations by ConvNets. In  addition, we will try to compare or combine this layer with other approaches such as deformable ConvNets \cite{dai-qi-xiong-li-2017} or depthwise ConvNets \cite{khan-nui-2021}, among others. Toward that goal,  further work is needed to compare the functionality of our approach to other methods and techniques such as object detection or segmentation.





\section*{Appendix}

\subsection*{Quaternion Algebra} \label{apen:quater}
The quaternion algebra $\mathbf{H}$ is a four dimensional real vector space with basis $1, \ii, \jj, \kk$,
\begin{equation}
    \mathbf{H} = \mathbf{R} 1 \oplus \mathbf{R} \ii  \oplus \mathbf{R} \jj   \oplus \mathbf{R} \kk 
\end{equation}
endowed with the bilinear product (multiplication) defined by Hamilton's relations, namely
\begin{equation}\label{eq:hamilton-rels-1}
\ii^2= \jj^2 = \kk^2 = \ii\jj\kk = -1.
\end{equation}
As it is easily seen, these relations imply that
\begin{equation}\label{eq:hamilton-rels-2}
\ii\jj=-\jj\ii=\kk,\quad \jj\kk=-\kk\jj=\ii, \quad \kk\ii=-\ii\kk = \jj.
\end{equation}
The elements of $\HH$ are named \emph{quaternions}, and $\ii,\jj,\kk$, \emph{quaternionic units}.
By definition, a quaternion $q$ can be written in a unique way in the form
\begin{equation}
q = a +b \ii + c \jj +d \kk,\quad a,b,c,d\in \mathbf{R}.
\end{equation}
Its \emph{conjugate}, $\bar{q}$, is defined as
\begin{equation}
\bar{q} = a -(b \ii + c \jj +d \kk).
\end{equation}
Note that $(q+\bar{q})/2=a$, which is called the \emph{real part}  or \emph{scalar part}  of $q$, and
$(q-\bar{q})/2=q-a=b\ii+c\jj+d\kk$, the \emph{vector part} of $q$.

Since the conjugates of $\ii,\jj,\kk$ are $-\ii,-\jj,-\kk$,
the relations \eqref{eq:hamilton-rels-1} and \eqref{eq:hamilton-rels-2}
imply that the conjugation is an \emph{antiautomorphism} of $\mathbf{H}$,
which means that it is a linear automorphism such that
$\overline{qq}'=\bar{q}'\bar{q}$. Using Hamilton's relations again, we easily conclude that
\begin{equation}
q\bar{q} = a^2+b^2+c^2+d^2.
\end{equation}
This allows to define the \emph{modulus} of $q$, $\abs{q}$, as the unique non-negative real number
such that
\begin{equation}
\abs{q}^2 = q\bar{q}.
\end{equation}
Observe that $\abs{qq'}=\abs{q}\abs{q'}$. Indeed,
$\abs{qq'}^2 = qq'\overline{qq}'=qq'\bar{q}'\bar{q}=q\abs{q'}^2\bar{q}=\abs{q}^2\abs{q'}^2$.

Finally, for $q\ne0$, $|q|>0$ and $q(\bar{q}/|q|^2)=1$, which shows that any non-zero quaternion has an inverse and therefore that $\mathbf{H}$ is a (skew) field.

\bibliographystyle{ieeetr}
\bibliography{ML,new_bibs}

\providecommand{\noopsort}[1]{}
\begin{thebibliography}{10}

\bibitem{rolls-stringer-2006}
E.~T. Rolls and S.~M. Stringer, ``Invariant visual object recognition: a model,
  with lighting invariance,'' {\em Journal of Physiology-Paris}, vol.~100,
  no.~1-3, pp.~43--62, 2006.

\bibitem{hendrycks-dietterich-2019}
D.~Hendrycks and T.~Dietterich, ``Benchmarking neural network robustness to
  common corruptions and perturbations,'' 2019.
\newblock \url{https://arxiv.org/pdf/1903.12261.pdf}.

\bibitem{dodge-karam-2016}
S.~Dodge and L.~Karam, ``Understanding how image quality affects deep neural
  networks,'' in {\em Quality of Multimedia Experience (QoMEX), 2016 Eighth
  International Conference on}, pp.~1--6, IEEE, 2016.

\bibitem{wang-yao-2018}
Y.~Wang, F.~Fu, F.~Lai, W.~Xu, J.~Shi, and J.~Wang, ``Efficient road specular
  reflection removal based on gradient properties,'' {\em Multimedia Tools and
  Applications}, vol.~77, no.~23, pp.~30615--30631, 2018.

\bibitem{halicek-febelo-fei-2019}
M.~Halicek, H.~Fabelo, S.~Ortega, J.~V. Little, X.~Wang, A.~Y. Chen, G.~M.
  Callico, L.~Myers, B.~D. Sumer, and B.~Fei, ``Hyperspectral imaging for head
  and neck cancer detection: specular glare and variance of the tumor margin in
  surgical specimens,'' {\em Journal of Medical Imaging}, vol.~6, no.~3,
  p.~035004, 2019.

\bibitem{simard-steinkraus-platt-2003}
P.~Y. Simard, D.~Steinkraus, and J.~C. Platt, ``Best practices for
  convolutional neural networks applied to visual document analysis,'' in {\em
  International Conference on Document Analysis and Recognition
  \emph{(ICDAR)}}, p.~958, IEEE Computer Society, 2003.

\bibitem{hernandez-konig-2018}
A.~Hern{\'a}ndez-Garc{\'\i}a and P.~K{\"o}nig, ``Data augmentation instead of
  explicit regularization,'' 2018.
\newblock \url{https://arxiv.org/pdf/1806.03852.pdf}.

\bibitem{cohen-welling-2016}
T.~Cohen and M.~Welling, ``Group equivariant convolutional networks,'' in {\em
  International conference on machine learning}, pp.~2990--2999, 2016.

\bibitem{ratner-ehrenberg-etal-2017}
A.~J. Ratner, H.~R. Ehrenberg, Z.~Hussain, J.~Dunnmon, and C.~R{\'e},
  ``Learning to compose domain-specific transformations for data
  augmentation,'' {\em Advances in neural information processing systems},
  vol.~30, p.~3239, 2017.

\bibitem{dao-gu-ratner-etal-2019}
T.~Dao, A.~Gu, A.~Ratner, V.~Smith, C.~De~Sa, and C.~R{\'e}, ``A kernel theory
  of modern data augmentation,'' in {\em International Conference on Machine
  Learning}, pp.~1528--1537, PMLR, 2019.

\bibitem{krizhevsky-sutskever-hinton-2012}
A.~Krizhevsky, I.~Sutskever, and G.~E. Hinton, ``{ImageNet classification with
  deep convolutional neural networks},'' in {\em Proceedings of the Conference
  Neural Information Processing Systems (NIPS 2011)}, pp.~1097--1105, 2012.
\newblock \url{http://www.image-net.org/}.

\bibitem{rad-roth-lepetit-2020}
M.~Rad, P.~M. Roth, and V.~Lepetit, ``{ALCN: Adaptive Local Contrast
  Normalization},'' {\em Computer Vision and Image Understanding}, vol.~194,
  p.~102947, 2020.

\bibitem{fan-yang-2017}
Q.~Fan, J.~Yang, G.~Hua, B.~Chen, and D.~Wipf, ``A generic deep architecture
  for single image reflection removal and image smoothing,'' in {\em
  Proceedings of the IEEE International Conference on Computer Vision},
  pp.~3238--3247, 2017.

\bibitem{hubel-wiesel-1962}
D.~H. Hubel and T.~N. Wiesel, ``Receptive fields, binocular interaction and
  functional architecture in the cat's visual cortex,'' {\em The Journal of
  physiology}, vol.~160, no.~1, pp.~106--154, 1962.

\bibitem{lecun-bottou-orr-muller-1998}
Y.~LeCun, L.~Bottou, G.~B. Orr, and K.-R. M{\"u}ller, ``Efficient backprop,''
  in {\em Neural networks: {T}ricks of the trade}, pp.~9--50, Springer, 1998.

\bibitem{han-kashif-2017}
H.~Xiao, K.~Rasul, and R.~Vollgraf, ``Fashion-mnist: a novel image dataset for
  benchmarking machine learning algorithms,'' {\em CoRR}, vol.~abs/1708.07747,
  2017.

\bibitem{krizhevsky-Hinton-etothers-2009}
A.~Krizhevsky, G.~Hinton, {\em et~al.}, ``Learning multiple layers of features
  from tiny images,'' 2009.
\newblock
  \url{http://citeseerx.ist.psu.edu/viewdoc/download?doi=10.1.1.222.9220&rep=rep1&type=pdf}.

\bibitem{elson-doucer-2007}
J.~Elson, J.~J. Douceur, J.~Howell, and J.~Saul, ``Asirra: A captcha that
  exploits interest-aligned manual image categorization,'' in {\em Proceedings
  of 14th ACM Conference on Computer and Communications Security (CCS)},
  Association for Computing Machinery, Inc., October 2007.

\bibitem{moya-xambo-perez-salazar-mzortega-cortes-2020}
E.~U. Moya-S{\'a}nchez, S.~Xamb{\'o}-Descamps, A.~S. P{\'e}rez,
  S.~Salazar-Colores, J.~Mart{\'\i}nez-Ortega, and U.~Cort{\'e}s, ``A
  bio-inspired quaternion local phase cnn layer with contrast invariance and
  linear sensitivity to rotation angles,'' {\em Pattern Recognition Letters},
  vol.~131, pp.~56--62, 2020.

\bibitem{leibo-liao-2015}
J.~Z. Leibo, Q.~Liao, F.~Anselmi, and T.~Poggio, ``The invariance hypothesis
  implies domain-specific regions in visual cortex,'' {\em PLoS Computational
  Biology}, vol.~11, no.~10, p.~e1004390, 2015.

\bibitem{worrall-garbin-daniyar-brostow-2017}
D.~E. Worrall, S.~J. Garbin, D.~Turmukhambetov, and G.~J. Brostow, ``Harmonic
  networks: Deep translation and rotation equivariance,'' in {\em Proceedings
  of the IEEE Conference on Computer Vision and Pattern Recognition},
  pp.~5028--5037, 2017.

\bibitem{richards-paiement-xie-duc-2020}
F.~Richards, A.~Paiement, X.~Xie, and P.-A. Duc, ``{Learnable Gabor modulated
  complex-valued networks for orientation robustness},'' 2020.
\newblock \url{https://arxiv.org/pdf/2011.11734.pdf}.

\bibitem{dapello-marques-schrimpf-geiger-cox-dicarlo-2020}
J.~Dapello, T.~Marques, M.~Schrimpf, F.~Geiger, D.~Cox, and J.~J. DiCarlo,
  ``Simulating a primary visual cortex at the front of {CNNs} improves
  robustness to image perturbations,'' {\em Advances in Neural Information
  Processing Systems}, vol.~33, 2020.

\bibitem{alshammari-akcay-breckon-2018}
N.~Alshammari, S.~Akcay, and T.~P. Breckon, ``On the impact of
  illumination-invariant image pre-transformation for contemporary automotive
  semantic scene understanding,'' in {\em 2018 IEEE Intelligent Vehicles
  Symposium (IV)}, pp.~1027--1032, IEEE, 2018.

\bibitem{moya-xambo-salazar-sanchez-cortes-2021}
E.~U. Moya-S\'anchez, S.~Xamb\'o-Descamps, S.~Salazar~Colores,
  A.~S\'anchez~P\'erez, , and U.~Cort\'es, ``{A Quaternion Deterministic
  Monogenic CNN Layer for Contrast Invariance},'' in {\em {Systems, Patterns
  and Data Engineering with Geometric Calculi}} (A.~Delshams and
  S.~Xambó-Descamps, eds.), ICIAM2019 SEMA SIMAI Springer Series,
  pp.~131--149, Springer, 2021.

\bibitem{xambo-2018-spins}
S.~Xamb\'o-Descamps, {\em {Real spinorial groups---a short mathematical
  introduction}}.
\newblock SBMA/Springerbrief, Springer, 2018.

\bibitem{smith-2007}
J.~O. Smith, {\em Mathematics of the discrete Fourier transform (DFT): with
  audio applications}.
\newblock Julius Smith, 2007.

\bibitem{kay-1978}
S.~{Kay}, ``Maximum entropy spectral estimation using the analytical signal,''
  {\em IEEE Transactions on Acoustics, Speech, and Signal Processing}, vol.~26,
  no.~5, pp.~467--469, 1978.

\bibitem{boukerroui-noble-2004}
D.~Boukerroui, J.~A. Noble, and M.~Brady, ``On the choice of band-pass
  quadrature filters,'' {\em Journal of Mathematical Imaging and Vision},
  vol.~21, no.~1-2, pp.~53--80, 2004.

\bibitem{granlund-knudsson-1995}
G.~H. Granlund and H.~Knutsson, {\em Signal processing for computer vision}.
\newblock Springer, 1995.

\bibitem{moya-santacruz-2011}
E.~U. Moya-S{\'a}nchez and E.~V{\'a}zquez-Santacruz, ``A geometric bio-inspired
  model for recognition of low-level structures,'' in {\em International
  Conference on Artificial Neural Networks}, pp.~429--436, Springer, 2011.

\bibitem{bayro-santacruz-moya-castillo-2016}
E.~Bayro-Corrochano, E.~Vazquez-Santacruz, E.~Moya-Sanchez, and
  E.~Castillo-Mu{\~n}is, ``{Geometric bioinspired networks for recognition of
  2-D and 3-D low-level structures and transformations},'' {\em IEEE
  transactions on neural networks and learning systems}, vol.~27, no.~10,
  pp.~2020--2034, 2016.

\bibitem{moya-bayro-2014}
E.~U. Moya-S{\'a}nchez and E.~Bayro-Corrochano, ``Symmetry feature extraction
  based on quaternionic local phase,'' {\em Advances in Applied Clifford
  Algebras}, vol.~24, no.~2, pp.~333--354, 2014.

\bibitem{felsberg-sommer-2001}
M.~Felsberg and G.~Sommer, ``The monogenic signal,'' {\em IEEE Transactions on
  Signal Processing}, vol.~49, no.~12, pp.~3136--3144, 2001.

\bibitem{bigun-2006}
J.~Bigun, {\em Vision with direction}.
\newblock Springer, 2006.

\bibitem{xing-yeh-2014}
D.~Xing, C.-I. Yeh, J.~Gordon, and R.~M. Shapley, ``Cortical brightness
  adaptation when darkness and brightness produce different dynamical states in
  the visual cortex,'' {\em Proceedings of the National Academy of Sciences
  USA}, vol.~111, no.~3, pp.~1210--1215, 2014.

\bibitem{daugman-1985}
J.~G. Daugman, ``{Uncertainty relation for resolution in space, spatial
  frequency, and orientation optimized by two-dimensional visual cortical
  filters},'' {\em J. Opt. Soc. Am. A}, vol.~2, pp.~1160--1169, jul 1985.

\bibitem{nava-escalante-cristobal-2012}
R.~Nava, B.~Escalante-Ram{\'\i}rez, and G.~Crist{\'o}bal, ``Texture image
  retrieval based on log-gabor features,'' in {\em Iberoamerican Congress on
  Pattern Recognition}, pp.~414--421, Springer, 2012.

\bibitem{smith-1978}
A.~R. Smith, ``Color gamut transform pairs,'' {\em ACM Siggraph Computer
  Graphics}, vol.~12, no.~3, pp.~12--19, 1978.

\bibitem{agoston-2005}
M.~K. Agoston, {\em Computer graphics and geometric modeling}, vol.~1.
\newblock Springer, 2005.

\bibitem{goodfellow-bengio-courville-2016}
I.~Goodfellow, Y.~Bengio, and A.~Courville, {\em Deep learning}.
\newblock MIT press, 2016.

\bibitem{lecun-cortes-burges-2010}
Y.~LeCun, C.~Cortes, and C.~Burges, ``{MNIST handwritten digit database},''
  {\em ATT Labs [Online]. Available: http://yann.lecun.com/exdb/mnist}, vol.~2,
  2010.

\bibitem{abadi-et39-2016}
M.~Abadi, A.~Agarwal, P.~Barham, E.~Brevdo, Z.~Chen, C.~Citro, G.~S. Corrado,
  A.~Davis, J.~Dean, M.~Devin, {\em et~al.}, ``Tensorflow: Large-scale machine
  learning on heterogeneous distributed systems,'' {\em arXiv preprint
  arXiv:1603.04467}, 2016.

\bibitem{McCartney-Hall-Freeman-1977}
E.~J. McCartney and F.~F. Hall, ``{Optics of the atmosphere: scattering by
  molecules and particles},'' {\em Physics Today}, vol.~30, no.~5, pp.~76--77,
  1977.

\bibitem{he-sun-tang-2011}
K.~He, J.~Sun, and X.~Tang, ``Single image haze removal using dark channel
  prior,'' {\em IEEE transactions on pattern analysis and machine
  intelligence}, vol.~33, no.~12, pp.~2341--2353, 2011.

\bibitem{he-mcauley-2016}
R.~He and J.~McAuley, ``{Ups and downs: Modeling the visual evolution of
  fashion trends with one-class collaborative filtering},'' in {\em Proceedings
  of the 25th International Conference on World Wide Web}, pp.~507--517,
  International World Wide Web, 2016.
\newblock \url{http://jmcauley.ucsd.edu/data/amazon/}.

\bibitem{chollet-et-al-2015}
F.~Chollet {\em et~al.}, ``Keras.'' \url{https://github.com/keras-team/keras},
  2015.

\bibitem{gao-rehman-wang-2011}
Y.~Gao, A.~Rehman, and Z.~Wang, ``Cw-ssim based image classification,'' in {\em
  2011 18th IEEE International Conference on Image Processing}, pp.~1249--1252,
  IEEE, 2011.

\bibitem{dai-qi-xiong-li-2017}
J.~Dai, H.~Qi, Y.~Xiong, Y.~Li, G.~Zhang, H.~Hu, and Y.~Wei, ``Deformable
  convolutional networks,'' in {\em Proceedings of the IEEE international
  conference on computer vision}, pp.~764--773, 2017.

\bibitem{khan-nui-2021}
Z.~Y. Khan and Z.~Niu, ``Cnn with depthwise separable convolutions and combined
  kernels for rating prediction,'' {\em Expert Systems with Applications},
  vol.~170, p.~114528, 2021.

\end{thebibliography}



\end{document}